\documentclass[sigconf]{acmart}
\usepackage{soul}
\usepackage{wrapfig} 
\usepackage{tabularx}
\usepackage{minted}
\usepackage{listings}
\usepackage{makecell}
\usepackage{subcaption}
\usepackage{xcolor}

\colorlet{punct}{red!60!black}
\definecolor{background}{HTML}{EEEEEE}
\definecolor{delim}{RGB}{20,105,176}
\colorlet{numb}{magenta!60!black}

\lstdefinelanguage{json}{
    basicstyle=\normalfont\ttfamily,
    numbers=left,
    numberstyle=\scriptsize,
    stepnumber=1,
    numbersep=8pt,
    showstringspaces=false,
    breaklines=true,
    frame=lines,
    backgroundcolor=\color{background},
    literate=
     *{0}{{{\color{numb}0}}}{1}
      {1}{{{\color{numb}1}}}{1}
      {2}{{{\color{numb}2}}}{1}
      {3}{{{\color{numb}3}}}{1}
      {4}{{{\color{numb}4}}}{1}
      {5}{{{\color{numb}5}}}{1}
      {6}{{{\color{numb}6}}}{1}
      {7}{{{\color{numb}7}}}{1}
      {8}{{{\color{numb}8}}}{1}
      {9}{{{\color{numb}9}}}{1}
      {:}{{{\color{punct}{:}}}}{1}
      {,}{{{\color{punct}{,}}}}{1}
      {\{}{{{\color{delim}{\{}}}}{1}
      {\}}{{{\color{delim}{\}}}}}{1}
      {[}{{{\color{delim}{[}}}}{1}
      {]}{{{\color{delim}{]}}}}{1},
}

\AtBeginDocument{%
  }

\begin{document}

\title{Autonomous Data Agents: A New Opportunity for Smart Data}

\author{Yanjie Fu}
\email{yanjie.fu@asu.edu} 
\affiliation{%
  \institution{Arizona State University}
  \state{Arizona}
  \country{USA}
}

\author{Dongjie Wang}
\email{wangdongjie@ku.edu} 
\affiliation{%
  \institution{University of Kansas}
  \state{Kansas}
  \country{USA}
}

\author{Wangyang Ying, Xinyuan Wang}
\email{wangyang.ying,xwang735@asu.edu} 
\affiliation{%
  \institution{Arizona State University}
  \state{Arizona}
  \country{USA}
}

\author{Xiangliang Zhang}
\email{xzhang33@nd.edu} 
\affiliation{%
  \institution{University of Notre Dame}
  \state{Illinois}
  \country{USA}
}

\author{Huan Liu}
\email{huanliu@asu.edu} 
\affiliation{%
  \institution{Arizona State University}
  \state{Arizona}
  \country{USA}
}

\author{Jian Pei}
\email{j.pei@duke.edu} 
\affiliation{%
  \institution{Duke University}
  \state{North Carolina}
  \country{USA}
}

\renewcommand{\shortauthors}{Yanjie et al.}
\begin{abstract}
As data continues to grow in scale and complexity, preparing, transforming, and analyzing it remains labor-intensive, repetitive, and difficult to scale. Since data contains knowledge and AI learns knowledge from it, the alignment between AI and data is essential. However, data is often not structured in ways that are optimal for AI utilization. Moreover, an important question arises: how much knowledge can we pack into data through intensive data operations? Autonomous data agents (DataAgents), which integrate LLM reasoning with task decomposition, action reasoning and grounding, and tool calling, can autonomously interpret data task descriptions, decompose tasks into subtasks, reason over actions, ground actions into python code or tool calling, and execute operations. Unlike traditional data management and engineering tools, DataAgents dynamically plan workflows, call powerful tools, and adapt to diverse data tasks at scale. This report argues that DataAgents represent a paradigm shift toward autonomous data-to-knowledge systems. DataAgents are capable of handling collection, integration, preprocessing,  selection, transformation, reweighing, augmentation, reprogramming, repairs, and retrieval. Through these capabilities, DataAgents transform complex and unstructured data into coherent and actionable knowledge. We first examine why the convergence of agentic AI and data-to-knowledge systems has emerged as a critical trend. We then define the concept of DataAgents and discuss their architectural design, training strategies, as well as the new skills and capabilities they enable.  Finally, we call for concerted efforts to advance action workflow optimization, establish open datasets and benchmark ecosystems, safeguard privacy, balance efficiency with scalability, and develop trustworthy DataAgent guardrails to prevent malicious actions. 
\end{abstract}
\maketitle
\section{Introduction}
\begin{figure}
\vspace{-0.1cm}
    \centering
    \includegraphics[width=1.0\linewidth]{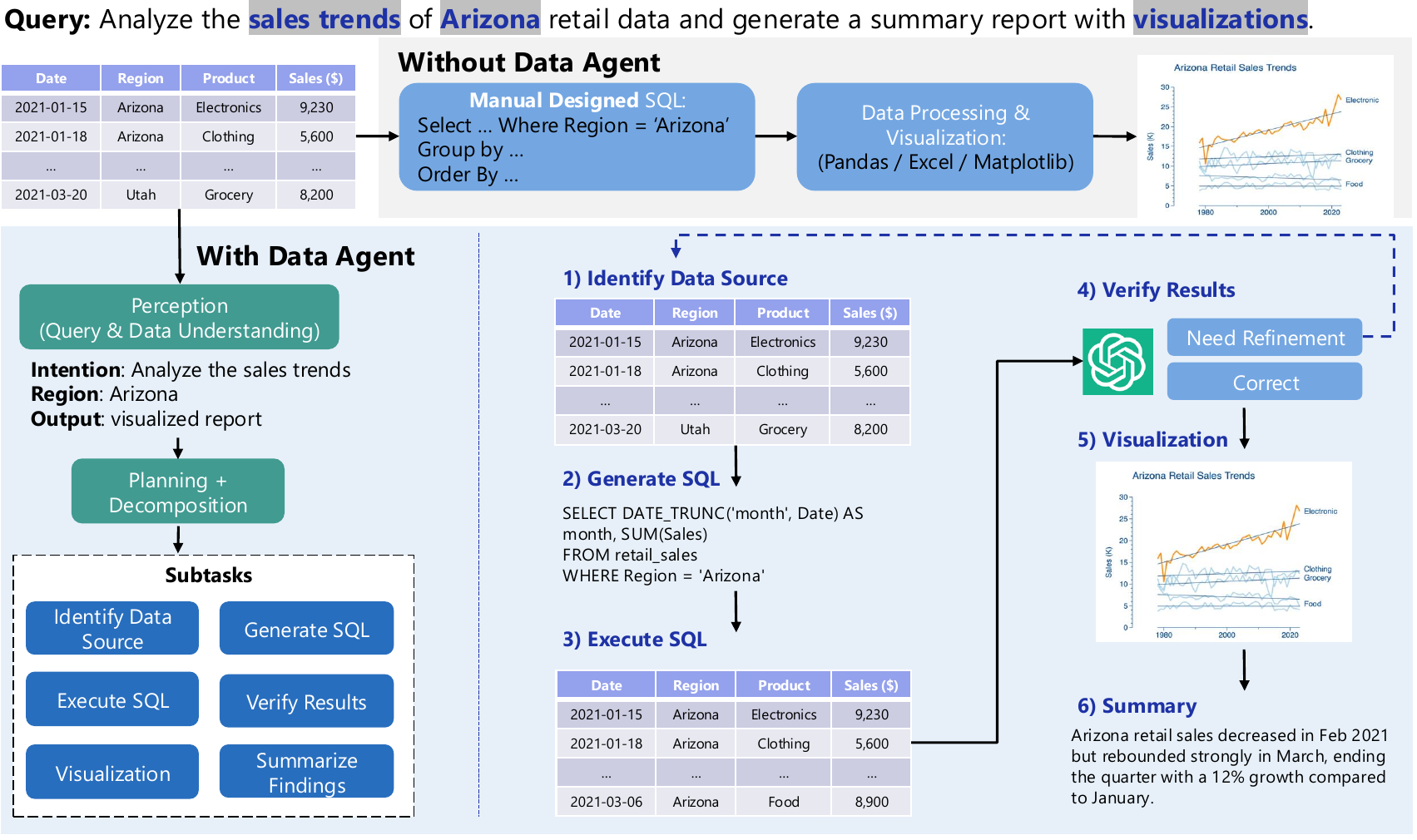}
    \caption{Data tasking without v.s. with DataAgents.}
    \label{fig:intro}
    \vspace{-0.7cm}
\end{figure}

As data become increasingly essential to AI, it has reshaped many data-driven tasks and applications like data collection, preprocessing, integration, feature engineering, data augmentation, equation extraction, text to SQL, data question answering, data retrieval, machine learning. 
For instance, feature engineering can leverage selection, transformation, and generation to create a new augmented feature set as data representation space, in order to develop discriminative data patterns for comparative machine learning. 
Despite the growing importance of quantity data in prediction and decision making, many of daily data-driven tasks remain repetitive, labor-intensive, and time-consuming~\textbf{(Figure \ref{fig:intro})}. 
For example, when preparing data for analysis, users often need to repeatedly conduct data cleaning (e.g., handle missing values, addressing outliers, remove duplicate records, correcting inconsistent data entries and data types), data transformation (e.g., standardization, normalization, discretizations, One-Hot encoding, label encoding), feature engineering (e.g., selection, transformation, generation). 
To perform these data tasks, one of the most promising approaches is to develop \textbf{autonomous data agents (DataAgents)} that \textit{equip data with the ability to take task description as inputs, understand tasks, decompose and plan tasks, reason actions, ground actions, execute actions to operate data, so that data can think, speak, and act}. 
DataAgents can reason, plan, ground, execute continuously without fatigue or loss of accuracy to extract knowledge from not just structure data, but also hybrid, multi-modal, semi-structure data, enhance analytical throughput, and accelerate knowledge discovery and decision-making.

Conventional data-related tasks, such as, data collection, integration, preprocess, engineering,  retrieval, and analysis, have relied heavily on manual efforts and existing tools, thus, require practitioners to construct SQL queries, write complex data handling scripts, and interpret intermediate outputs. 
Although human expertise is effective, this manual or semi-manual process is time-consuming, error-prone, and difficult to scale. 
Recently,  advances in reinforcement learning have enabled partial automation of these tasks~\cite{fan2020autofs,fan2021interactive,wang2023reinforcement,xiao2023traceable,wang2022group}. For instance, reinforcement intelligence can learn optimal strategies for querying, transforming, and analyzing data under a targeted task~\cite{muller2022towards,abdelaal2024reclean}.
More recently, generative AI and large language models (LLMs) have demonstrated their broad world knowledge, structured data understanding, and instruction-following proficiency~\cite{gong2025evolutionary,gong2025unsupervised,gong2025neuro,abhyankar2025llm,hollmann2023large,ko2025ferg,liu2024survey,nguyen2024interpretable}. This exhibits a strong potential to perform complex data tasks based on user instructions. 
For instance,  the study in \cite{ying2024feature} regards feature selection operations as tokens, and formulate feature selection, originally a discrete search task, into a generative learning task with continuous optimization in embedding space under an encoding-optimization-decoding framework. 

Unlike traditional solutions that require manual coding or empirical experiences, the emergence of agentic AI offers a promising new direction: by combining data with LLM reasoning, planning, tool usage, reflection, DataAgents can be instructed by a task description to independently retrieve datasets, clean records, generate SQL queries, analyze results, and present summaries without minimum user intervention~\cite{kojima2022large,schick2023toolformer,peng2023instruction}. 
\textbf{Figure \ref{fig:intro}} shows that users can simply issue a natural language instruction, such as “Analyze the sales trends of Arizona retail data from Q1 2021 and generate a summary report with visualizations.” 
The DataAgent can then locate the relevant dataset, perform necessary preprocessing, run appropriate analyses, generate plots, and output a comprehensive report. In this way, we automate the entire analytic pipeline and improving workflow efficiency. 
In parallel, WebAgents~\cite{ning2025survey} automate web-related tasks, but there is increasing interest in transferring the WebAgent concept to data to automate data-related tasks for better data knowledgization. 

To bridge this gap, we provide a comprehensive overview of DataAgents by summarizing representative methods from the perspectives of architecture, training, applications, and challenges.  
Specifically, we first introduce the background knowledge of traditional agents and the fundamental pipeline of DataAgents. 
We then review the data perception, task planning, action space, action reasoning, memory management, action grounding, and action execution of data agents.  
We summarize two crucial aspects (i.e., training data and training strategies) in DataAgents training.  
Finally, we discuss promising future research directions in DataAgents.

\vspace{-0.4cm}
\section{Background of Autonomous Data Agents}

\noindent{\bf Data-related Tasks.}
Data-driven applications rely on diverse data tasks, including data collection, engineering, retrieval,  analysis, and visualization. 
Specifically, data engineering involves data cleaning,  data preprocessing, feature extraction, feature selection, feature transformation, feature generation. 
Data retrieval focuses on identifying and extracting relevant information from databases, APIs, and other storage systems, thus, require complex query formulation and schema understanding. 
Data analysis includes exploratory data analysis, predictive modeling, and visualization. 
These tasks are labor-intensive and require domain-specific skills and tools.

\noindent{\bf The Need and Feasibility of Automating Data Tasks.}
The data tasks, like data selection, transformation, generation, retrieval, prediction, summary, QA, visualization,  are repetitive and time-consuming. 
For instance, analysts often rewrite similar SQL queries across projects, manually clean and reconcile messy datasets, or repeatedly configure dashboards for recurring reports. 
Automating data tasks can improve data quality, make data AI-ready, reduce human effort, increase data usability, and accelerate time-to-insight. 
Like self-driving vehicles learn to perceive and act in physical environments, many data tasks share the logical procedural structure of data workflows, thus, people can code and execute the logical procedures for repetitive uses. 

\noindent{\bf Reinforcement Learning and Generative AI for Automating Data Engineering, Retrieval, and Analysis} 
To enable automation, recent efforts have explored reinforcement learning and generative AI for data tasks. 

Many data tasks can be framed as a decision process. Reinforcement learning can learn optimal policies for selecting features, transforming data, making decisions to perform feature crossing, or composing data pipelines, by interacting with the environment and receiving feedback. 
As one example, reinforcement agents have been trained to optimize query plans~\cite{xiong2024autoquo,tzoumas2008reinforcement,heitz2019join}, detect anomalies~\cite{arshad2022deep,muller2022towards}, and automate data cleaning steps~\cite{abdelaal2024reclean,peng2024rlclean}. 
As another example, researchers have formulated self optimizing feature selection, transformation, and generation as decision sequences generated by reinforcement policy networks~\cite{liu2021automated,fan2020autofs,fan2021interactive,liu2019automating,liu2021efficient,he2025fastft,liu2021efficient,wang2022group}. 

In parallel, many data tasks can be seen as a sequence of data operations or actions, for instance, $f_1*f_2, f_1/f_2, f_2/f_3, \sqrt{f4}$ as a generated token sequence. These operations can be seen as natural language or a sequence of special tokens. 
Generative AI models have demonstrated strong abilities in learning patterns from natural  languages and generate token responses. 
Similarly, generative AI models can be trained to learn from these data operations and corresponding performance scores from verifiers, to generate what actions we should take (executable python code, SQL queries, or direct operations) to complete data tasks to maximize data operations utilities. 
For instance, researchers have developed generative feature selection~\cite{gong2025neuro,ying2024feature,xiao2023beyond,gong2025neuro}, generative feature transformation and generation~\cite{wang2023reinforcement,xiao2023traceable,ying2023self,ying2024unsupervised}, generative data programming~\cite{bai2025privacy}. 

However, these generative models directly generate final answers at one step, thus, ignoring the power of many existing tools for data selection, transformation, and augmentation. Besides, there is no reasoning across intermediate steps to identify the high-confidence candidate solutions. For complex data tasks, generative models either require a huge depth of neural networks, or fail to generate valid and effective data operations. 

\noindent{\bf LLMs for Data Tasks.}
Recently, LLMs transforms this landscape. 
LLMs exhibit advanced reasoning and language comprehension, and can interact and adapt to external environments and verifier feedback. 
Researchers have generalized LLMs to translate ambiguous instructions from humans into concrete data operations and integrate with data tools to address feature selection~\cite{gong2025agentic}, feature transformation~\cite{gong2025unsupervised}, feature engineering~\cite{abhyankar2025llm, ko2025ferg,hollmann2023large}, data generation~\cite{hsieh2025dall}, texts to SQL and data retrieval~\cite{liu2024survey}, tabular data QA~\cite{zhu2024tat,zhu2024autotqa, nguyen2024interpretable}, data labeling~\cite{wang2024human,wang2024human,lu2025llm,zhang2025efficient}, data cleaning~\cite{biester2024llmclean,li2024autodcworkflow,naeem2024retclean,zhang2024data}, data preprocessing~\cite{zhang2023large,meguellati2025large}. 
\emph{LLMs can reason and generate texts given a query in a reactive fashion via single-turn interactions, however, LLMs are not goal-driven systems in an autonomous fashion, and, thus, cannot plan, decompose, reason, and act to accomplish tasks. }

\vspace{-0.3cm}
\section{Framing Autonomous Data Agents}
DataAgents are a new class of autonomous tasking systems (i.e., LLM + planner + reasoning + tools, instead of just a generative model) designed to interact with data as active problem-solvers rather than passive tools. 
Different from LLMs (e.g., GPT model + prompt), DataAgents integrate LLMs' language understanding, along with task planning and decomposition, action reasoning, tool calling,  action grounding and execution capabilities~\cite{schick2023toolformer,longpre2023flan,peng2023instruction,wang2024survey}. 
DataAgents can understand the high-level intent from the description of a data task (i.e., a query), decompose the task into multiple subtasks, reason action sequences for each subtask, execute data workflows, interact with external environments, reflect and adapt to changes. 
For instance, when given an instruction such as, “Identify trends in patient admissions and suggest resource allocation for next month,” a DataAgent can access hospital records, clean and structure the data, run time-series analyses, and generate a set of actionable recommendations without human inputs.

\textbf{Formally}, given the data (e.g., a table) and a user instruction (e.g., ``Identify trends in patient admissions and suggest resource allocation for next month''), 
DataAgents will generate a sequence of executable actions ``$A=\{a_1, a_2, a_3, ...a_n \}$'' and interact with the environment based on these actions to complete the user-instructed data task. 
Specifically, at the step $t$, DataAgents first observe the current environmental information (e.g., schema, features, instances, size and shape, data distribution) $s_t$ from the data, and retrieve the previous actions $(a_1, a_2, ..., a_{t-1})$ as the short-term memory to guide the next-action prediction. 
After that, the user instruction $q$, observations $s_t$, and previous actions $(a_1, a_2, ..., a_{t-1})$, are combined as the in-context knowledge to generate the next action $a_t$:
$$a_t = f(q, s_t, (a_1, a_2, ..., a_{t-1}))$$
where $f$ is an LLM based data agent. 
Finally, DataAgents will interact with and operate the data environment based on the generated operation $a_t$. 
The state of the data will be updated to $s_{t+1}$ after executing the generated action, defined as:
$$s_{t+1}= S(a|t)$$
DataAgents will iteratively repeat the aforementioned steps until the user-given data task is completed.

\textbf{Illustrative Example.}  
Unlike LLMs that only provide static responses, DataAgents operate through a closed-loop pipeline that integrates perception, planning, reasoning, execution, and refinement. 
Consider the hospital management task: ``Predict ICU bed demand for next month and suggest resource allocation.'' 
A vanilla LLM may only output general statements such as ``admissions may increase in winter,'' but it cannot directly perceive database schemas, decompose the task, or run executable analyses. 
In contrast, a DataAgent first \emph{perceives} the structure and distribution of admission data, then \emph{plans and decomposes} the task into subtasks (data cleaning, forecasting, attribution). 
Next, it \emph{reasons about the action sequence}, \emph{grounds} each action into executable queries or API calls, and \emph{executes} them on the hospital data system. 
Finally, it performs \emph{refinement} by incorporating feedback from intermediate results to adjust the workflow and improve accuracy. 
This step-by-step autonomy demonstrates why DataAgents are necessary: they enable adaptive, end-to-end problem solving in real data environments, which cannot be achieved by prompting an LLM alone.

\section{DataAgents Architecture}
Figure 2 shows the important components of designing DataAgents for autonomous data tasking: 
1) \textbf{Perception}:  requires DataAgents to observe and quantify the current data task environment (e.g., data task description, targeted data). 
2) \textbf{Planning \& Decomposition}: requires DataAgents to understand and analyze an user-given task descriptions based on the current environment, decompose the data task into a logical structure of subtasks (e.g., sequential, parallel, selection), and reason corresponding actions to solve each subtask. 
3) \textbf{Grounding \& Execution}: requires DataAgents to translate actions into summaries, code, or tool calling, execute the generated actions, and interact with the environment. 
In the following, we will review the important techniques employed by DataAgents in these steps.

\begin{figure}
    \centering
    \includegraphics[width=1.0\linewidth]{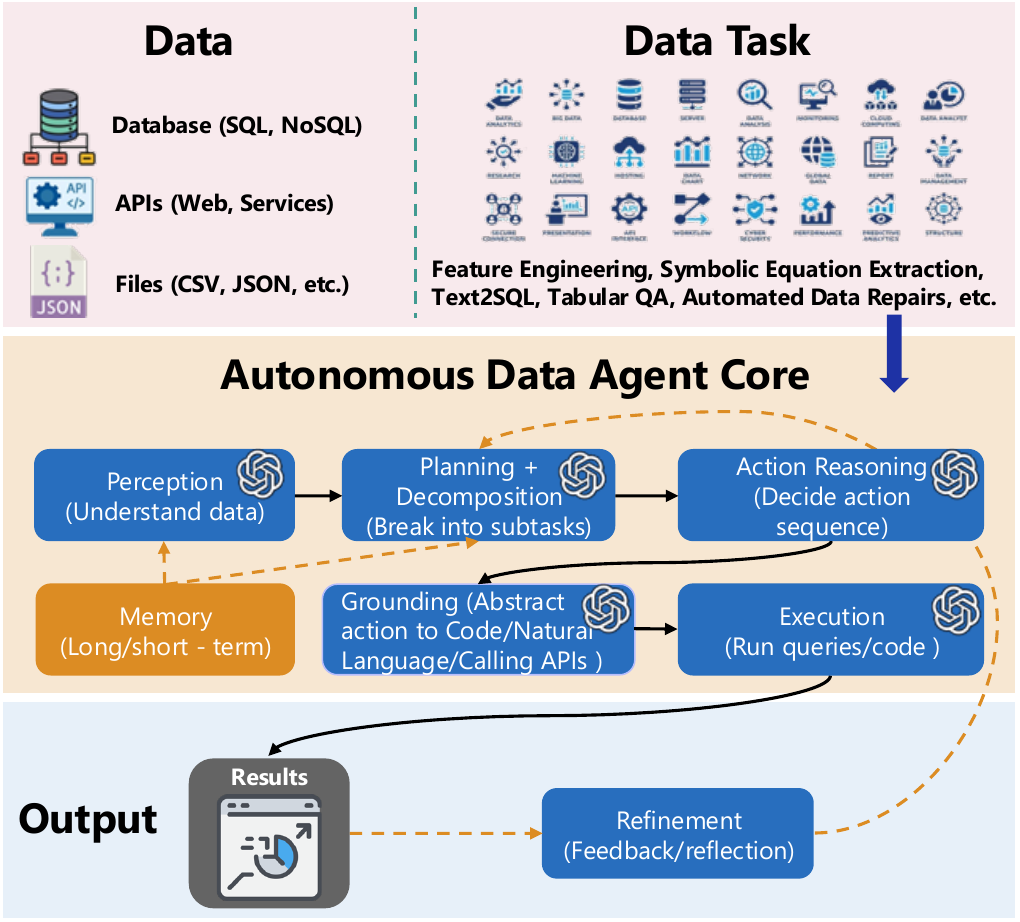}
    \caption{Framework of Data Agents}
    \label{fig:enter-label}
\end{figure}

\vspace{-0.2cm}
\subsection{Perception: Representing Task Descriptions and Data Contexts}
DataAgents operating within complex data (e.g., tables, time series, sequences, graphs and networks, spatial data, events, texts, images, videos, audio) should perceive accurate data environment, understand task description, and perform task decomposition and reasoning based on data environments combined with the given task.
When given a task description and target data, DataAgents need to (i) analyze the structure (e.g., schema), content (e.g., value distributions), and context (e.g., time range, units, domain) of the dataset, (ii) extract tasking intents, constraints, and expectations from the natural language prompt of task descriptions~\cite{shojaee2024llm}. 
For example, when given a prompt like “clean this hospital admission data for trend analysis,” DataAgents must infer and understand that the missing values in timestamps should be imputed,  the categorical variables like diagnosis codes need normalization, and temporal patterns require sorting or aggregation. These inferences need the agent’s ability to link task description with concrete data structures and operations. A task (prompt) is usually described in texts, while target data can be in different modalities, thus, we can propose different types of DataAgents:

\noindent\textbf{Data as tokens:}
LLMs excel in handling natural language, thus DataAgents usually leverage the textual representation of target data (e.g., table as texts, time series as texts, sequences as texts) to perceive the data environment.  

\noindent\textbf{Data as tokens and visualization:}
VLMs have enhanced agents in processing images and videos. To leverage the visual understanding capabilities of VLMs, we can integrate VLMs into DataAgents to utilize not just tokens but also visualizations to perceive the data patterns and environment. 
 
\noindent\textbf{Data as multi-modal semi-structure information box: } 
We propose to use a JSON structure to unify the representation of multi-modalities. 
Listing \ref{lst:json-example} shows that a JSON format not just store structured data like tables, time series, sequences, graphs, but also include
unstructured data like natural languages, code in the plain text blocks, as well as store images using URLs. 

\begin{lstlisting}[caption=Any Data as JASON, language=json,firstnumber=1, label=lst:json-example]
{
  "title": "Albert Einstein",
  "infobox": {
    "Born": "14 March 1879",
    "Died": "18 April 1955",
    "Fields": ["Theoretical physics"],
    "Known for": ["Theory of relativity", "Photoelectric effect"]
  },
  "text": "Albert Einstein was a German-born theoretical physicist...",
  "images": [
    {
      "url": "https://upload.wikimedia.org/example.jpg",
      "caption": "Einstein in 1921"
    }
  ]
}}
\end{lstlisting}

\subsection{Task Planning and Decomposition: from Data Task to Multiple Subtasks}

Task planning and decomposition involves dividing a high-level data task or query into multiple smaller, logical subtasks that are more manageable and executable. 
This decomposition enables agents to handle complex data workflows by breaking down them into multiple steps, such as, data retrieval, cleaning, analysis, visualization, and validation. 
In this way, we allow for iterative execution, error handling, and integration with tools including databases or APIs. 
For instance, a task like "analyze sales data for trends" might be decomposed into multiple subtasks: i) query database for raw data, ii) preprocess outliers, iii) apply statistical models, iv) generate visualizations, and v) interpret results.
Solutions for task decomposition in LLM-based agents typically leverage prompting techniques, search methods, and hierarchical structures to break down tasks into subtasks. Specifically, there are three solutions:
\noindent\textbf{ 1) CoT Prompting Based Task Decomposition} prompts the LLM to generate step-by-step reasoning traces.
It decomposes tasks into sequential subtasks through natural language chains.
This approach enhances logical breakdown for data analysis and also reduces errors in multi-step processes.
\noindent\textbf{ 2) Recursive Task Decomposition}, where the agent recursively breaks down task and subtasks into finer-grained ones until they are atomic and executable, often using programmatic or LLM-driven recursion, ideal for handling nested data queries or complex hierarchies.
\noindent\textbf{ 3) Tree-based Search Task Decomposition},  which explores a tree of possible decompositions via branching thoughts or actions, selecting optimal subtask paths through search algorithms like Monte Carlo Tree Search, effective for exploratory data tasks with multiple pathways.

Table \ref{table:task_decomposition} show that, given a task: ``From our e-commerce data warehouse, produce a weekly revenue trend for the past quarter, segmented by product category, and flag anomalies where weekly revenue drops by more than 20\% compared to the prior week.'', DataAgents need to 
1) parse textual task description (inputs) and ouput structured task specifications (goal, inputs, outputs, and constraints); 
2) load columns, types, and relationships; 
4) select relevant time range and fields via SQL;
5) join and enrich transactions with product categories; 
6) aggregate weekly revenue; 
7) detect anomalies — compare weekly revenue vs. prior week; 
8) generate report — table + chart + anomaly highlights.

{\footnotesize
\begin{table}[h]
\caption{Decomposing A Task into Subtasks}
\vspace{-0.1cm}
\begin{tabularx}{\linewidth}{|X|X|X|X|}\hline
\label{table:task_decomposition}
Subtask & Input & Output & Dependency \\
\hline
Parse Task & Natural language task description. & Structured specs: goal, inputs, outputs, constraints. & None \\
\hline
Load Schema & E-commerce database connection. & Schema: columns, types, relationships. & Parsed specs \\
\hline
Plan Query & Schema, time range, fields. & SQL query plan. & Schema \\
\hline
Select Data & Query plan. & Raw data: date, revenue, product\_id, category. & Query plan \\
\hline
Join Data & Raw data. & Enriched dataset with categories. & Selected data \\
\hline
Aggregate Revenue & Enriched dataset. & Weekly revenue by category. & Joined data \\
\hline
Detect Anomalies & Weekly revenue data. & Flagged >20\% drops. & Aggregated data \\
\hline
Generate Report & Data with anomalies. & Table, chart, anomaly highlights. & Anomalies \\
\hline
\end{tabularx}
\end{table}
}

Task decomposition methods can be improved from the three aspects:
1) \textit{decomposition strategies}: creating recursive, adaptive methods that allow agents to decompose tasks on-the-fly to generate task graphs or trees for complex, multi-hop queries while incorporating feedback loops for refinement in uncertain data scenarios; 
2) \textit{cost efficiency}: exploring ensembles of fine-tuned, smaller models for decomposition to reduce computational costs and improve affordability, particularly for data-intensive tasks, while maintaining performance comparable to larger LLMs.
3) \textit{safety}: understanding how decomposition can bypass safety guardrails in LLMs and developing aligned methods that ensure decomposed subtasks remain safe, verifiable, and resistant to misuse in sensitive data applications.

\subsection{Action Reasoning: from Subtask to Action Sequence}

\subsubsection{Action Space}
The actions of DataAgents are three fold: tool calling (e.g., invoke APIs, database engine,  python data preprocessing tools, python feature engineering libs), symbolic expressions (e.g., produce structured formal queries, code, or transformation specs), direct generation (e.g., output natural language summaries, insights, or computations in-model). This action space serves as the set of all possible “moves” the DataAgents can make to fulfill a data task.

\noindent\textbf{Tool calling as actions.}
A tool calling action is defined as the action that the agent prepare valid inputs (e.g., arguments) and invoke an external tool (e.g., pandas, sklearn, or SQL engines) to complete the work and doesn’t directly produce the result itself, but issues commands~\cite{schick2023toolformer}. 
For instance, a data agent can call a database connector to run a SQL query and fetch data, call a data profiling tool (e.g., Pandas Profiling, Great Expectations) for data quality metrics,  call a machine learning toolkit (e.g., scikit-learn, TensorFlow) to train or evaluate a model, or call a plotting library (e.g., Matplotlib, Plotly) to generate a chart from query results.
Tool calling introduces constraints (e.g., versioning, resource access), so agents must reason about action feasibility and fallback options.

\textbf{Symbolic expressions as actions.}
A symbolic expression action is defined as when the agent generates  formal symbolic declarative representations (e.g.,  structured formal queries, code, or transformation specs) that are then executed by some systems~\cite{kamienny2022end}. 
One example is to generate SQL for filtering, joining, aggregating, or sampling data: ``SELECT customer\_id, SUM(sales) FROM orders  WHERE order\_date >= '2025-01-01' GROUP BY customer\_id;''. 
Another example is to generate Pandas/PySpark transformation code. We next discuss symbolic expression actions in two essential data tasks: feature selection and generation.

\underline{\it Data task 1: feature selection.} 
Feature selection is a common symbolic task where the agent chooses a subset of informative features based on statistical metrics or model-based criteria. 
Symbolic actions include expressions like \( \text{select}(f_i \,|\, \text{MI}(f_i, y) > \theta) \) or \( \text{rank}(f_i, \text{importance}) \). 
The agent evaluates feature importance using mutual information, correlation coefficients, or model weights~\cite{abhyankar2025llm,hollmann2023large}.

\underline{\it Data task 2: feature transformation.}
In feature transformation, agents modify raw features to create discriminative patterns. Symbolic actions include normalization \( z = \frac{x - \mu}{\sigma} \), log scaling \( x' = \log(x + 1) \), or binning \( \text{bin}(x, 5) \). These expressions can be automatically generated and applied based on feature distribution characteristics.

\underline{\it Data task 3: feature generation.}
Feature generation involves constructing new variables by crossing existing ones. Symbolic generation includes interaction terms \( f_1 \cdot f_2 \), ratios \( \frac{f_1}{f_2} \), or non-linear combinations \( \sqrt{f_3} + \log(f_4) \). Agents can learn feature crossing patterns, and generate optimal feature transformations via optimization.

\noindent\textbf{Direct generation as actions.}
When the agent directly produces the end output without intermediate tool invocation — the result is computed, reasoned, or drafted entirely within the LLM or AI runtime~\cite{zhu2024large}.
One example is to summarize dataset insights in natural language from already-provided data, like: ``Sales in Q1 increased by 15\% compared to Q4, mainly driven by product category A.''
Another example is responding to analytical questions using only the visible dataset, like: Q: “Which product had the highest sales in March?” A: “Product B with 3,240 units sold.”

Generally speaking, simple tasks are executed symbolically, while complex ones leverage external tools. 

\vspace{-0.2cm}
\subsubsection{Guiding Next Action Generation via Memory Management}
Memory management refers to how DataAgents with LLM as reasoning core keep track of context and past interactions within a high-level task and across subtasks. 
Memory mechanisms can help data agents to maintain contextual awareness and sequence consistency.  
There are two types of memories, including short-term and long-term memories, that can enable agents to trace historical actions and anticipate next actions~\cite{sun2018memory}.

\underline{\it (1) Short-term Memory: Recent Actions.}
 Short-term memory is the most recent actions within a single subtask. For example, after encoding a categorical feature, the agent can use short-term memory to avoid re-encoding or applying incompatible feature transformations.  Short-term memory is subtask-specific and cleared when task is completed in order to optimize agents for reactive task handling.
 
\underline{\it (2) Long-term Memory: Action Trajectories.}
Long-term memory is historical action trajectories across tasks. Similar to reinforcement learning, it allows agents to learn the appropriateness of an action for a task type. For instance, if an outlier detection method is effective on financial datasets, long-term memory recommends  this method strategy in similar tasks.

Memory management can be improved by: 
1) memory representation and retrieval optimization to lower retrieval latency while maximizing relevance tp retrieve relevant past interactions or facts;
2) memory selection and forgetting, by developing algorithms to decide what memory should be kept, summarized, or discarded; 3) integration of short-term and long-term memory for coherent multi-turn reasoning. 

\subsubsection{Action Reasoning.} 
Once a complex task is broken down into logical subtasks (e.g., data querying, analysis, visualization), action reasoning combines planning, search, and reinforcement learning principles, and usually involves iterative cycles of thinking about the subtask, selecting appropriate actions (such as tool calls or code generation), observing outcomes, and refining based on feedback. 
There are three major categories of action reasoning methods:
 
\ul{\it (1) Reasoning and Acting (ReAct)}, which interleaves reasoning (generating thoughts about a target subtask) with action execution and observation, in order to create iterative loops to refine sequences until completion. It's widely used for dynamic subtasks requiring tool integration, like sequential data queries. Recent research has been focused on self-reflection mechanisms (e.g., linguistic feedback) to adjust action sequences based on real-time observations, reducing errors and improving success rates in long-horizon subtasks like data analysis chains.

\ul{\it (2) Chain-of-Thought (CoT)}: By prompting the data agent to build step-by-step reasoning trajectories, CoT formalizes action sequences as logical chains, to enhance performance in symbolic or mathematical subtasks such as data computation.

\ul{\it (3) Monte Carlo Tree Search}: A lookahead search algorithm that simulates potential action sequences, balancing exploration and exploitation to select optimal paths. It's ideal for uncertain subtasks, like optimizing data processing paths in complex environments.

Action-action dependency is critical. For instance, in a subtask requiring outlier removal and normalization, the agent must reason to remove outliers first to avoid distorting normalization. Such reasoning ensures logical coherence and robustness in data workflows~\cite{wei2021finetuned}.

\vspace{-0.2cm}
\subsection{Action Grounding and Execution}

\subsubsection{Action Grounding.}
action grounding refers to the process of mapping an abstract action (derived from task decomposition and reasoning) to concrete, executable operations in the real-world environment or tools. This involves translating an abstract symbolic or generated action into specific, grounded actions like generating SQL queries, calling APIs, executing Python code for data processing, or interacting with databases.  The agent must ensure that variable names, data types, and function signatures are all matched to their execution context. Without proper grounding, even a syntactically correct action may fail during runtime. We summarize three types of action grounding strategies:

\ul{\it (1) Ground an action to executable code}: which involves translating an agent's action into executable code, typically Python, that is then run in an interpreter to interact with the environment. Today's LLM can generate code snippets that can include control flows (e.g., loops, conditionals), data manipulation, and tool integrations. For instance, the CodeAct consolidates agent actions into executable code (e.g., Python) as a unified space, enabling grounding of diverse tasks like data analysis by generating and running code snippets, with built-in error handling and observation.

\ul{\it (2) Ground an action to structured tool invocations}: which invokes predefined tools or APIs through structured formats like JSON-based tool calling. The LLM outputs a specific call (e.g., tool name and arguments), which is parsed and executed externally. 

\ul{\it (3) Ground an action to direct natural language generation}, which grounds actions directly in natural language outputs generated by the LLM, without external execution or tools. The agent produces text responses, explanations, or decisions as the "action," often for user interaction or simple reasoning tasks. For instance, LLM-Planner~\cite{song2023llm}, is a few-shot grounding approach that leverages LLMs to generate executable plans and actions from natural language, focusing on embodied or data tasks with minimal examples for efficient execution.

\subsubsection{ Action Execution.} Execution entails the actual invocation of these actions, observation of outcomes (e.g., query results or errors), and iterative refinement or error handling to complete subtasks. This phase often incorporates feedback loops, such as reflection on results or replanning, to adapt to dynamic data environments and mitigate issues like hallucinations or invalid operations.

For instance, after decomposing a task into subtasks like "retrieve data" and "compute statistics," grounding might convert "retrieve data" to a precise database API call, and execution would run it, process the output, and pass it to the next subtask.

\begin{table*}[!ht]
\centering
\caption{Data sources and training instances for multi-skill data AI agents}
\label{tab:data_ai_agent_tasks}
\begin{tabular}{|p{3cm}|p{5cm}|p{7cm}|}
\hline
\textbf{Task} & \textbf{Data Sources} & \textbf{Example Training Instance (Instruction $\rightarrow$ Input $\rightarrow$ Output)} \\ \hline
Data Preprocessing & Public noisy datasets (Kaggle, UCI), open-source cleaning scripts & ``Remove rows with missing age'' $\rightarrow$ Raw CSV $\rightarrow$ Cleaned CSV + code snippet \\ \hline
Feature Engineering & Kaggle notebooks, FEAST tutorials, synthetic tabular data & ``Create an indicator if income $>$ 50K'' $\rightarrow$ Table $\rightarrow$ New table + feature code \\ \hline
Data Augmentation & NLP: backtranslation corpora; CV: Albumentations scripts; Tabular: SMOTE datasets & ``Add Gaussian noise to the `price' column'' $\rightarrow$ Table $\rightarrow$ Augmented table + code \\ \hline
Data Visualization & Matplotlib/Plotly examples, ChartQA datasets & ``Plot sales over time'' $\rightarrow$ Table $\rightarrow$ Chart image + plotting code \\ \hline
Text-to-SQL & Spider, WikiSQL, real SQL logs with schema & ``List average salary by department'' $\rightarrow$ Schema $\rightarrow$ SQL query \\ \hline
Data-to-Equation & Symbolic regression benchmarks (SRBench), physics datasets & ``Fit a linear model for $y$ vs. $x$'' $\rightarrow$ CSV $\rightarrow$ Equation + coefficients \\ \hline
Tool Calling & Logs from Jupyter, SQL IDEs, BI dashboards, CLI usage & ``Summarize column `revenue' statistics'' $\rightarrow$ Table $\rightarrow$ Tool call JSON + expected output \\ \hline
\end{tabular}
\end{table*}

\section{DataAgents Training}

\subsection{Preliminaries of Instruction Tuning}
As a type of Supervised FineTuning (SFT), instruction tuning aims to align pre-trained LLMs with diverse data operation tasks by conditioning them on task instructions~\cite{peng2023instruction,longpre2023flan}. 
The idea is to teach LLMs to be prompted by natural language inputs and generate structured multi-step action sequences or executable operations for tasking. 
This process involves adjusting the model's internal parameters to better align its behavior with the given instructions. 
Instruction tuning can train agents with basic capabilities for understanding instructions, decomposing tasks, and producing preliminary action sequences. However, to achieve reliable multi-step planning, tool usage, and environment-adaptive execution, reinforcement-based fine-tuning (RFT) is necessary to further enhance the reasoning and operational capabilities of the agents.



\subsection{Preparing Training Data for Skills and Action Reasoning}

To train DataAgents with diverse skills (e.g., data preprocessing, feature engineering, data augmentation, data visualization, text-to-SQL translation,  data-to-equation mapping), it is essential to construct a multi-task, instruction-oriented dataset grounded in real-world workflows. 
\textbf{Table \ref{tab:data_ai_agent_tasks}} shows that the training dataset should adopt an instruction–input–output format, in order to allow agents to learn from natural language commands and corresponding execution traces.

\noindent\textbf{Skill 1: Data Preprocessing Training Data.} For data preprocessing, training samples can be derived from public datasets from Kaggle or the UCI Machine Learning Repository containing noise, missing values, duplicates, or heterogeneous formats, paired with human-curated cleaning results and executable code snippets. 

\noindent\textbf{Skill 2: Feature Engineering Training Data.} Feature engineering examples may be sourced from open repositories, competition notebooks, and synthetic datasets, along with code covering feature creation, transformation, and selection. 
    
\noindent\textbf{Skill 3: Data Augmentation Training Data.}  Data augmentation samples should span textual (e.g., back translation, synonym replacement), visual (e.g., geometric or color-space transformations), and tabular (e.g., SMOTE, noise injection) modalities. 

\noindent\textbf{Skill 4: Data Visualization Training Data.}  For data visualization, examples can be obtained from public plotting libraries and annotated chart-generation tasks, linking raw data to generated visualizations and their underlying code.  

\noindent\textbf{Skill 5: Text2SQL Training Data.}  Text-to-SQL training data can be drawn from established benchmarks such as Spider and WikiSQL, as well as domain-specific SQL query logs, each paired with schema definitions and validated queries.

\noindent\textbf{Skill 6: Data2Equation Training Data.} Data-to-equation examples can be generated from physics, engineering, and regression datasets with known analytical forms, or sourced from symbolic regression benchmarks, for instance, SRBench.

\noindent\textbf{Skill 7: Tool Calling Training Samples.}  Tool calling examples (\textbf{Listing \ref{lst:tool_calling_example}}) should pair natural language requests with tool metadata, parameterized invocation formats, and corresponding execution results. Such datasets can be sourced from instrumented Jupyter notebooks, business intelligence platforms, or query-generation interfaces, where both the intermediate tool calls and final outputs are logged.

\begin{lstlisting}[language=json, caption={Example of a tool calling training instance}, label={lst:tool_calling_example}]
Instruction: "Draw a histogram of the 'age' column."
Input: CSV schema: name, age, salary
Tool Call: {
  "tool": "matplotlib.hist",
  "args": {
    "data": "age",
    "bins": 10
  }
}
Output: <image file or base64-encoded chart>
\end{lstlisting}

To ensure broad generalization, data collection should combine multiple strategies: mining open-source code repositories and competition platforms for high-quality task–solution pairs; generating synthetic datasets with controlled transformations to establish precise ground truth; crowdsourcing expert annotations for complex or ambiguous tasks; and logging human–tool interactions in authentic analytical workflows.

\subsection{Preparing Training Data for Task Decomposition}
Collecting training data for task decomposition requires curating multi-stage examples in which a complex data analysis request is broken down into a logical sequence of heterogeneous subtasks such as data preprocessing, feature engineering, data augmentation, data visualization, text-to-SQL translation, data-to-equation mapping, and tool calling. 
Each training instance should pair the original high-level instruction with an explicit decomposition plan specifying the ordered plan of subtasks, the descriptions of each subtask, their dependencies, and the corresponding inputs and outputs, enabling the agent to learn both structural reasoning and skill invocation. \textbf{Listing \ref{lst:decomp_json}} shows an example task decomposition from a task description to multiple subtasks to actions. 
Such data can be sourced from annotated Jupyter notebooks, BI dashboards, competition notebooks, and open-source repositories that explicitly document intermediate steps, inputs, and outputs for each stage. 

\begin{lstlisting}[language=json, caption={An example of  task decomposition}, label={lst:decomp_json}]
{
  "complex_instruction": "Analyze customer purchase data to find seasonal trends and forecast next quarter's sales.",
  "plan": [
    {
      "step": 1,
      "skill": "preprocessing",
      "instruction": "Remove duplicates and standardize date formats.",
      "input": "raw_transactions.csv (columns: id, date, amount, category)",
      "output": "clean_transactions.csv; Python code used; artifact hash"
    },
    {
      "step": 2,
      "skill": "feature_engineering",
      "instruction": "Create month, year, and holiday flag features.",
      "input": "clean_transactions.csv",
      "output": "features.csv; code snippet; preview head"
    },
    {
      "step": 3,
      "skill": "augmentation",
      "instruction": "Synthesize data for missing holiday periods.",
      "input": "features.csv",
      "output": "augmented_features.csv; generation parameters"
    },
    {
      "step": 4,
      "skill": "visualization",
      "instruction": "Plot monthly sales trends for the last 3 years.",
      "input": "aggregated table: month, sales",
      "output": "matplotlib code; rendered chart (image ref)"
    },
    {
      "step": 5,
      "skill": "text_to_sql",
      "instruction": "Aggregate sales by month and category from the database.",
      "input": "schema: sales(order_id, date, amount, category)",
      "output": "SQL query; execution result snippet"
    },
    {
      "step": 6,
      "skill": "data_to_equation",
      "instruction": "Fit a seasonal model to forecast next quarter.",
      "input": "monthly sales time series",
      "output": "equation form + coefficients; R^2/MAE"
    },
    {
      "step": 7,
      "skill": "tool_call",
      "instruction": "Export forecast results to CSV.",
      "input": "forecasted series",
      "output": {
        "tool": "file.export_csv",
        "args": {"path": "forecast_q_next.csv", "data": "forecast_table"}
      }
    }
  ]
}
\end{lstlisting}

\begin{figure}[h]
    \centering
    \includegraphics[width=1.0\linewidth]{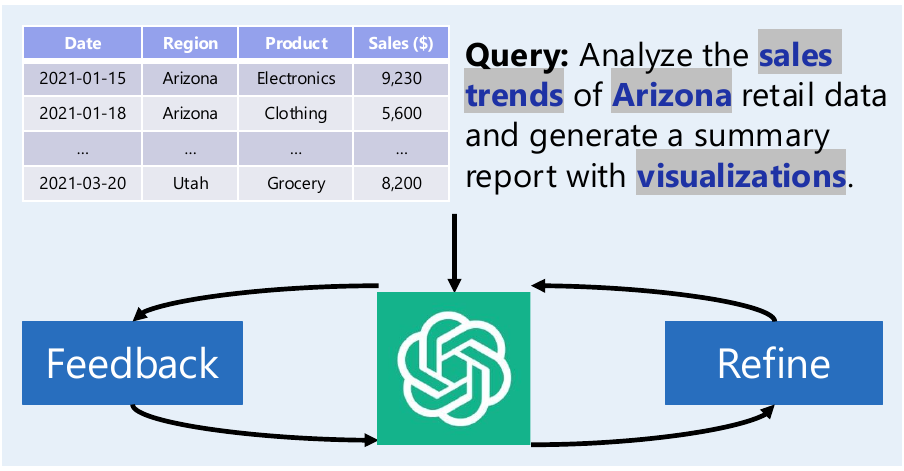}
    \caption{Single agent design.}
    \label{fig:single-agent}
    \vspace{-0.3cm}
\end{figure}

\vspace{-0.2cm}
\subsection{Single Agent Based Design and Instruction Tuning}
If we aim to train a single DataAgent to decompose, reason, and execute a complex data task, we can adopt a unified supervised instruction tuning approach. 
Each training instance includes the high-level task description, contextual information (e.g., schema snippets, table samples, constraints), and a complete task-subtask-action–execution trace. 
The trace begins with a structured plan in JSON format (\textbf{Listing \ref{lst:decomp_json}}) that details an ordered sequence of subtasks with associated skill types (e.g., preprocessing, feature engineering, data augmentation, data visualization, text-to-SQL, data-to-equation, tool calling) and step-specific instructions. 
This is followed by a sequence of action blocks, in which the agent produces grounded, executable outputs such as Python code, SQL queries, tool call JSON objects, or equations, and observation blocks containing deterministic execution results (e.g., preview tables, chart hashes, error logs). 
We can finetune the model end-to-end on these concatenated action traces using next-token cross-entropy loss, optionally with structure-aware loss weighting to improve the syntactic correctness of JSON, SQL, and code outputs. We can collect training data from verified multi-step task decompositions, to ensure that each action–observation pair is executable in a controlled sandbox environment, and include both successful and failure-recovery examples to promote robustness. 
At inference time, the same model iteratively generates a plan and then executes each step, incorporating tool execution feedback into subsequent actions, thereby enabling closed-loop reasoning and grounded multi-skill execution within a single generative architecture.

\begin{figure}[h]
    \centering
    \includegraphics[width=1.0\linewidth]{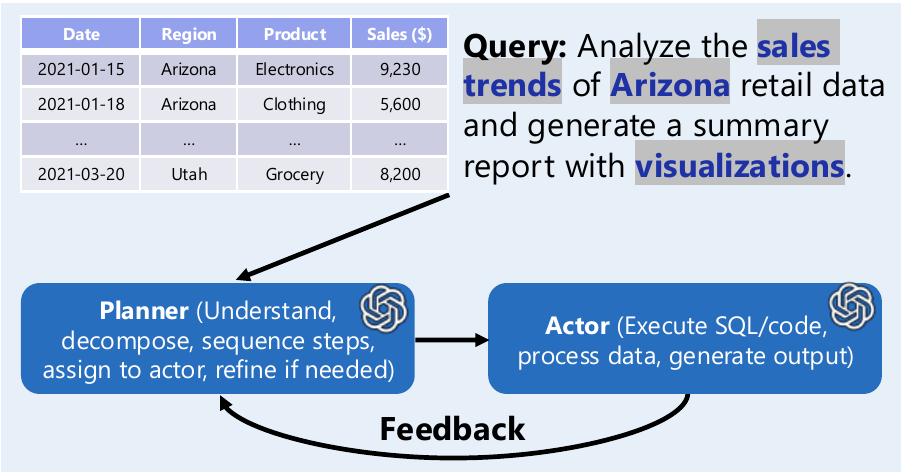}
    \caption{Planner-actor dual agent design.}
    \label{fig:dual_agents}
\end{figure}
\subsection{Planner-Actor Dual Agents Based Design and Instruction Tuning}
Besides, we can adopt a two-module training paradigm comprising a planner and an actor,  supervised on the task decomposition and subtask action datasets respectively. 
We can train the planner module to map a high-level natural-language instruction to an ordered sequence of subtasks, each annotated with its skill type (e.g., preprocessing, feature engineering, data augmentation, data visualization, text-to-SQL, data-to-equation mapping, tool calling) and a concise sub-instruction. 
We can train the actor module to condition on a single subtask specification and its associated input context (e.g., table snippet, database schema, chart specification) to produce the corresponding executable output, such as code, SQL, structured tool call JSON, or fitted equations. 
Both modules are fine-tuned on curated and synthetic multi-skill workflows, with the planner optimized using sequence modeling loss over serialized decomposition plans and the actor optimized using next-token prediction loss over serialized subtask outputs. 
In this way, we allow the agent to generalize decomposition strategies to novel tasks while maintaining high fidelity in tool-grounded action generation, in order to enable robust end-to-end performance in heterogeneous, multi-step data tasks.
Multiple multi-agent paradigms could be applied in data agents (e.g., role-specialized agents, hierarchical controllers, or critic-based supervision). We select the planner–actor design, as it offers conceptual clarity and aligns naturally with data-centric workflows, where task decomposition and reliable execution are the two most critical components. Other multi-agent designs can also be incorporated into the data-agent framework, but we use this paradigm mainly to illustrate the idea clearly.

\subsection{Reinforcement Fine-Tuning of DataAgents}

\begin{figure}[!ht]
\vspace{-0.2cm}
    \centering
    \includegraphics[width=1.0\linewidth]{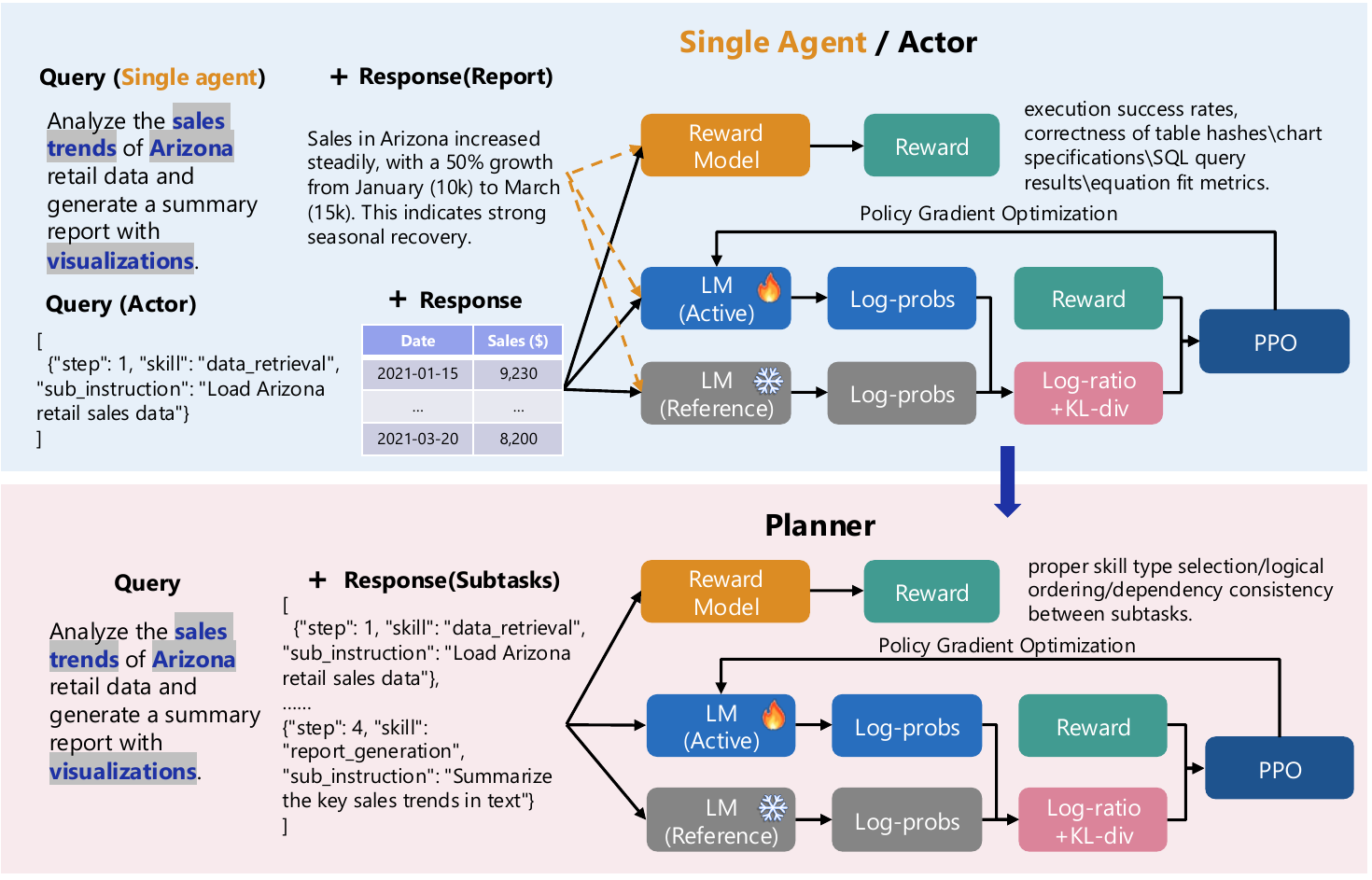}
    \caption{Reinforcement finetuning of LLM generative agent}
    \label{fig:rft}
\vspace{-0.3cm}
\end{figure}

\textbf{RFT Framework.} To further improve the DataAgents’ ability to perform accurate task decomposition and coherent action reasoning beyond supervised instruction tuning, we can employ reinforcement fine-tuning (\textbf{Figure \ref{fig:rft}}).
If this is a single agent design, we can exploit a single reinforcement policy as generative agent. If this is a planner-actor dual agent design,  we can exploit a hierarchical, two-policy setting with one policy as a task planner and the other policy as an subtask actor.

\textbf{Reward Model.}  DataAgents generate a decomposition plan and corresponding subtask actions for a given high-level data task, and each output is evaluated using a composite reward function that captures both structural and execution quality. 
For the planning stage, rewards are assigned based on the correctness and completeness of the subtask sequence, including proper skill type selection, logical ordering, and dependency consistency between subtasks. 
For the action reasoning stage, the action sequence of a subtask output is executed in a sandboxed environment, and rewards are computed from execution success rates, correctness of final artifacts (e.g., table hashes, chart specifications, SQL query results, equation fit metrics), and adherence to tool schema constraints. 

\textbf{Policy Optimization.} Reinforcement optimization, implemented via policy gradient methods such as Proximal Policy Optimization (PPO), iteratively updates the model to maximize these rewards, enabling it to refine both the decomposition structure and the reasoning behind action selection. This approach allows the agent to learn from real or simulated execution feedback, improving its robustness, adaptability, and end-to-end success in solving complex, multi-step analytical tasks.

\section{A Study of DataAgents}
\subsection{Primary Workflow}
\label{sec:workflow}
\begin{figure}[h]
  \centering
  \includegraphics[width=\linewidth]{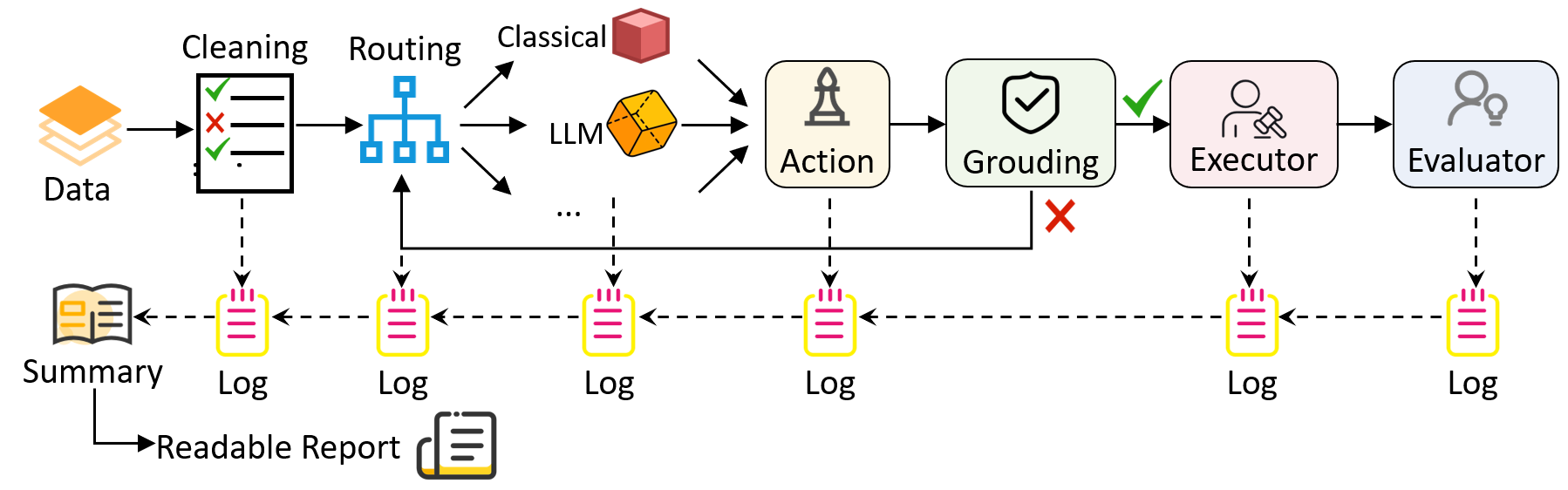}
  \caption{Overall workflow of the proposed DataAgent, consisting of sequential modules for data cleaning, routing, planning/action, grounding, execution, and evaluation.}
  \label{fig:workflow}
\end{figure}

\textbf{Overview.}
We evaluate a modular DataAgent that executes tabular ML tasks via a stepwise pipeline:
\emph{(i) cleaning / preprocessing} using a small rule-based controller (if--else rules over schema/profiled stats);\,
\emph{(ii) routing} by a large LLM (e.g., GPT family) from a natural-language dataset card to one of three solution modes: classical linear/sparse model (e.g., Lasso~\cite{tibshirani1996regression}), neural model (RL trained), or LLM-based sequence generator;\,
\emph{(iii) planning + tool calling} (merged): for classical/neural, it calls a fixed tool chain to shape a better feature set; for LLM mode, it generates a sequence of operations (plan-as-sequence~\cite{yao2023react,schick2023toolformer}) and calls tools accordingly;\,
\emph{(iv) grounding} that validates the called tools or the generated operation sequence (schema checks, dry-run, unit tests);\,
\emph{(v) execute} to produce downstream predictions/plots;\,
\emph{(vi) evaluator} that performs automatic sanity checks and triggers at most $K$ self-repair rounds~\cite{shinn2023reflexion,madaan2023self,gu2024survey}.
\emph{(vii) summary} from the log of each step and generate a readable report for non-LLM-experts.
\textbf{Figure~\ref{fig:workflow}} summarizes the overall workflow.

\paragraph{Why this design.}
The primary workflow is structured to mirror the evaluation dimensions of our study. 
Each module corresponds to a potential failure mode in end-to-end data analysis. Routing tests whether LLMs can align their modeling capacity with dataset characteristics. Planning and tool-calling probe the compositional reasoning ability needed to orchestrate operations. Grounding and evaluator explicitly validate and repair intermediate outputs to ensure robustness. 
The workflow serves as both a pipeline and a testbed, enabling controlled evaluation of each module’s impact.

\subsection{Experimental Design: Comparing Paradigms}

\noindent\textbf{Purpose.}
Our goal is to use experiments as a lens to understand the strengths and weaknesses of three major paradigms for tabular data engineering:  
(1) \emph{classical and AutoML baselines},  
(2) \emph{pure LLM approaches}, and  
(3) \emph{RL-based methods}.  
 \textit{DataAgent} serves as a notable example of emerging \emph{modular agent paradigm}, showing how agentic design balances reliability, autonomy, and efficiency.


\noindent\textbf{Datasets.}
We include classification tasks from \textit{UCIrvine} and regression tasks from \textit{OpenML}, covering both discrete and continuous prediction settings.

\noindent\textbf{Evaluation Aspects.}  
We examine four aspects that highlight the trade-offs among paradigms:  
(i) \textit{Predictive quality}, measured by 1-RAE for regression and F1-score for classification;  
(ii) \textit{Reliability}, measured by task success rate and number of attempts needed;  
(iii) \textit{Automation}, reflected in grounding error rates;  
(iv) \textit{Efficiency}, measured by time and calls required.



\subsection{Observations from Comparative Results}
\label{sec:improvements}

\noindent\textbf{Predictive Quality Across Paradigms.}  
Table~\ref{tab:downstream} summarizes results across regression and classification datasets.  
Classical and AutoML baselines achieve solid but limited performance.  
Pure LLM approaches show strong results on some datasets, but suffer from instability across tasks.  
RL-based policies can reach high accuracy, yet often at the expense of extensive training.  
In contrast, 
\textit{DataAgent}, as a modular agent instance, consistently delivers robust accuracy while avoiding training overhead, reflecting the potential of this paradigm for practical deployment in realistic scenarios.

\begin{figure}[thbp]
  \centering
  \begin{subfigure}[t]{0.48\linewidth}
    \centering
    \includegraphics[width=\linewidth]{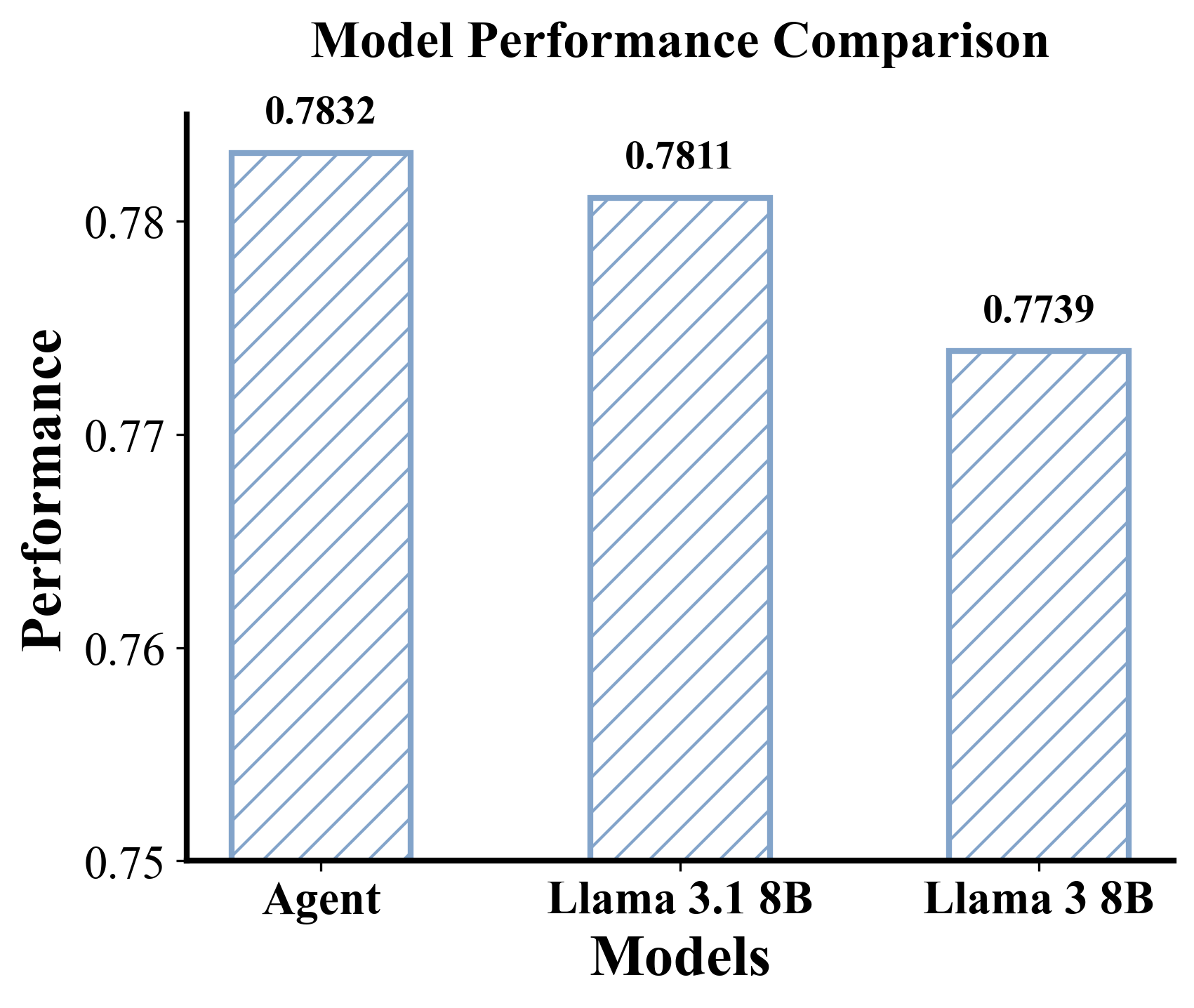}
    \caption{F1-score comparison.}
    \label{fig:autonomous_f1}
  \end{subfigure}
  \hfill
  \begin{subfigure}[t]{0.48\linewidth}
    \centering
    \includegraphics[width=\linewidth]{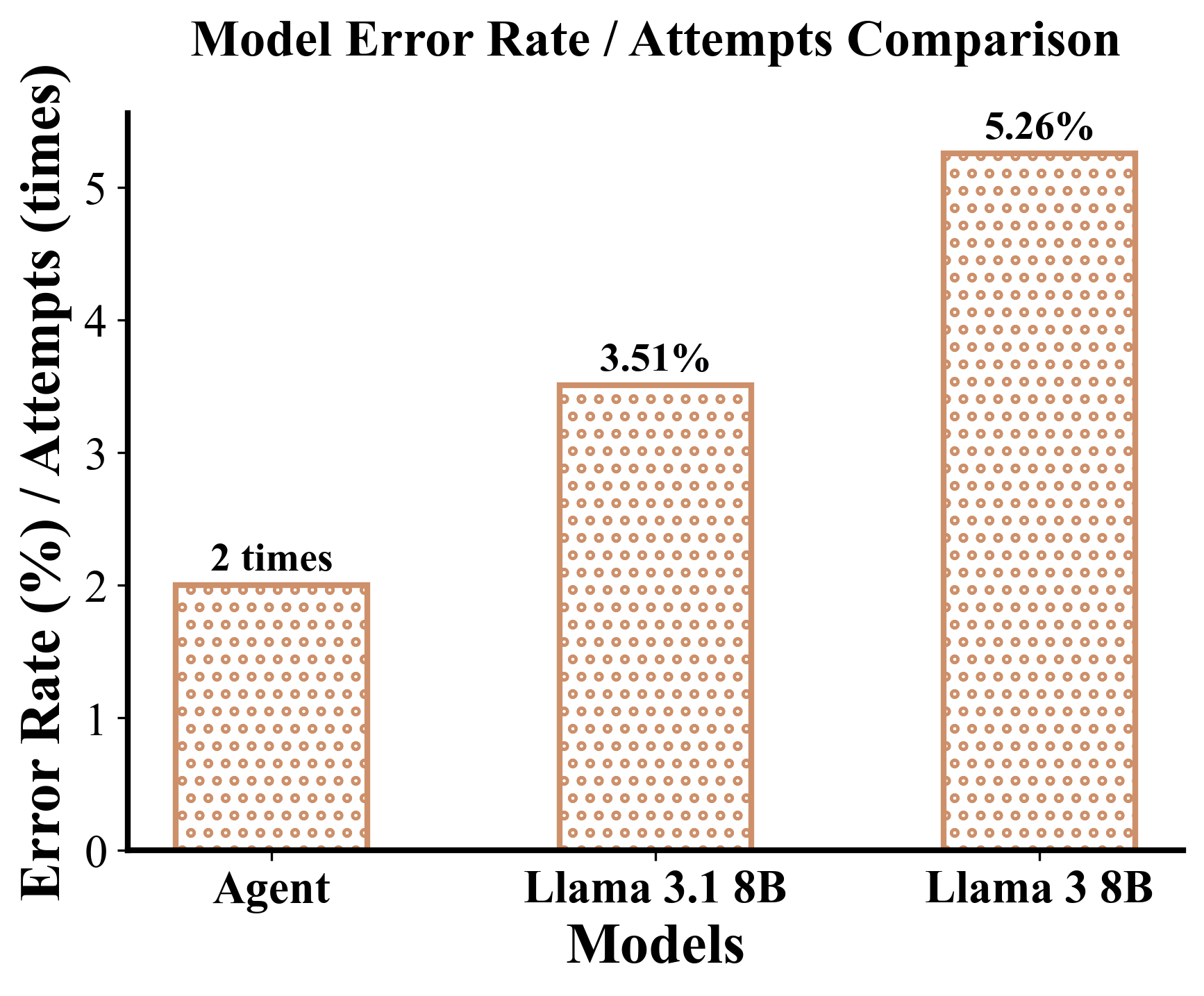}
    \caption{Error rate comparison.}
    \label{fig:autonomous_error}
  \end{subfigure}
  \caption{Comparison of modular agent (DataAgent) and pure LLM baselines, showing higher reliability and fewer retries for the modular agent paradigm.}
  \label{fig:autonomous}
\end{figure}

\noindent\textbf{Reliability and Automation.} 
Figure~\ref{fig:autonomous} compares DataAgent with pure LLM baselines on task performance and error rates.  
Pure LLMs often require multiple retries to produce an executable solution and exhibit higher grounding error rates.  
In contrast, \textit{DataAgent} achieves reliable task completion within a very small attempt budget (only 2 attempts on average), while maintaining the highest performance.  
This demonstrates how the modular agent design combines the reasoning ability of LLMs with structured grounding, ensuring both higher reliability and stronger automation.

\noindent\textbf{Efficiency Trade-offs}
Table~\ref{tab:efficiency} compares only RL-based policies with the modular agent (\textit{DataAgent}).
RL-Policy-1 and RL-Policy-2 incur substantial training costs and require many calls during inference.
\textit{DataAgent} is \emph{training-free}, reaches valid solutions with only \emph{2 calls}, and attains the strongest performance, despite a modest per-run inference latency.
This shows that RL-based methods require heavy training, whereas the modular agent avoids training and remains efficient with only light inference overhead.




\begin{table}[thbp]
  \begin{center}
    \caption{Downstream performance comparison on regression and classification datasets. 
    For regression, we report 1-RAE (Relative Absolute Error), and for classification F1-score (\%).}
    \label{tab:downstream}
    \resizebox{\linewidth}{!}{\begin{tabular}{lcccc}
        \toprule
        \textbf{Dataset} & German Credit & SpectF & OpenML 586 & OpenML 618 \\
        \midrule
        \textbf{Source} & UCIrvine~\cite{uci_dataset_2023} & UCIrvine~\cite{uci_dataset_2023} & OpenML~\cite{Openml_dataset_2023} & OpenML~\cite{Openml_dataset_2023} \\
        \textbf{Task} & Classification & Classification & Regression & Regression \\
        \textbf{Samples} & 1{,}000 & 267 & 1{,}000 & 1{,}000 \\
        \textbf{Features} & 24 & 44 & 25 & 50 \\
        \midrule
        \textbf{Original} & 74.20\% & 76.06\% & 0.6311 & 0.4402 \\
        \textbf{RDG} & 68.01\% & 76.03\% & 0.5681 & 0.3720 \\
        \textbf{LDA~\cite{blei2003latent}} & 63.91\% & 66.29\% & 0.1109 & 0.0521 \\
        \textbf{ERG} & 74.43\% & 75.66\% & 0.6147 & 0.3561 \\
        \textbf{NFS~\cite{chen2019neural}} & 68.67\% & 79.40\% & 0.5443 & 0.3473 \\
        \textbf{AFAT~\cite{horn2020autofeat}} & 68.32\% & 76.03\% & 0.5435 & 0.2472 \\
        \textbf{PCA~\cite{mackiewicz1993principal}} & 67.92\% & 70.92\% & 0.1109 & 0.1016 \\
        \textbf{TTG~\cite{khurana2018feature}} & 64.51\% & 76.03\% & 0.5443 & 0.3467 \\
        \textbf{FeatLLM~\cite{han2024large}} & 76.35\% & 80.07\% & 0.6477 & 0.4597 \\
        \textbf{CAAFE~\cite{hollmann2023large}} & 59.92\% & 70.60\% & N/A & N/A \\
        \textbf{AutoFeat~\cite{horn2019autofeat}} & 74.86\% & 76.06\% & 0.6329 & 0.4407 \\
        \textbf{OpenFE~\cite{zhang2023openfe}} & 74.50\% & 76.06\% & 0.6311 & 0.4402 \\
        \textbf{ELLM-FT~\cite{gong2025evolutionary}} & 76.39\% & 86.14\% & 0.6328 & 0.4734 \\
        \midrule
        \textbf{Pure LLM (Llama 3.1)} & 75.43\% & 85.19\% & 0.7811 & 0.6653 \\
        \textbf{RL-Policy-1 (GRFG~\cite{wang2022group})} & 68.29\% & 81.65\% & 0.5768 & 0.4562 \\
        \textbf{RL-Policy-2 (MOAT~\cite{wang2023reinforcement})} & 72.44\% & \textbf{86.95\%} & 0.6251 & 0.4734 \\
        \textbf{DataAgent} & \textbf{79.60\%} & 86.79\% & \textbf{0.7832} & \textbf{0.7488} \\
        \bottomrule
    \end{tabular}}
  \end{center}
\end{table}



\begin{table}[h]
  \begin{center}
    \caption{Efficiency comparison: time (seconds) and call times.}
    \label{tab:efficiency}
    \resizebox{\linewidth}{!}{\begin{tabular}{lccc}
        \toprule
        Metric & \makecell{RL-Policy-1 \\ (GRFG~\cite{wang2022group})} & \makecell{RL-Policy-2 \\ (MOAT~\cite{wang2023reinforcement})} & \makecell{DataAgent} \\
        \midrule
        Training Time (sec) & 1493.7 & 649.4 & 0 \\
        Inference Time (sec) & 0.5 & 2.14 & 3.89 \\
        Calls & 13 & 22 & 2 \\
        Performance & 0.5768 & 0.6251 & \textbf{0.7832} \\
        \bottomrule
    \end{tabular}}
  \end{center}
\end{table}

\subsection{Takeaways on Paradigms for Tabular Data}
\label{sec:insights}
From these comparisons, several insights emerge:

\paragraph{(1) Classical and AutoML baselines.}  
These methods are mature, stable, and widely used in practice.  
They provide a strong reference point, but their reliance on fixed model classes and handcrafted transformations limits adaptability to diverse and evolving datasets.  
As a result, they often plateau in performance when facing complex feature interactions.

\paragraph{(2) Pure LLM approaches.}  
LLMs show strong potential for automation by directly generating transformation or modeling sequences.  
However, their lack of grounding makes them prone to invalid outputs and silent failures, requiring multiple retries to succeed.  
This makes them less reliable as standalone solutions, though their flexibility remains valuable.

\paragraph{(3) RL-based policies.}  
RL methods can explore transformation spaces thoroughly and sometimes reach strong performance.  
Yet, they depend on heavy training for each dataset and require detailed environment-specific feedback loops.  
This investment in time and compute makes them less practical for broad deployment, despite their robustness.

\paragraph{(4) Modular agent paradigm (DataAgent).}  
The modular agent design strikes a balance across these trade-offs.  
It leverages LLMs for reasoning and exploration, but grounds outputs through explicit modules, ensuring task completion.  
Unlike RL, it is training-free and adapts zero-shot to new datasets, yet still retains exploratory diversity.  
This positions modular agents as a promising and practical paradigm for general-purpose tabular ML.

\section{DataAgents Capabilities}
Our perspective is that attaching agentic AI to data can equip the data with actionable capabilities, including autonomous feature engineering, symbolic equation extraction, text to sql, tabular QA, data quality assessment, data repairs, and more.

\vspace{-0.2cm}
\subsection{Automated Feature Engineering Capability} 
Positioned between data collection and machine learning, DataAgents leverage genAI, LLM, reasoning, tool calling to automate data-related tasks. 
These agents perceive data environments, reason about the space of instance, feature, target variables and multimodal data forms, and act to complete data preprocessing, integration, transformation, repairs, augmentation to streamline feature engineering. 
Taking feature selection as an example, DataAgents can: 
1) reason and call tools of statistical relevance tests (e.g., mutual information, ANOVA) and model-based methods (e.g., L1-regularized regression, gradient boosting feature importance) to remove irrelevant or redundant variables; 
2)regard feature selection as generating decision tokens and learn to generate feature selection decisions via instruction tuning from the training data of feature selection/disselection actions paired with corresponding downstream predictive accuracies. 
There are more studies related to generative and LLM-based feature selection, transformation, and generation studies, which can be used to finetune feature engineering-specific DataAgents~\cite{liu2021automated,wang2022group,xiao2023traceable,xiao2023traceable,xiao2023beyond,ying2023self,wang2023reinforcement,ying2024unsupervised,ying2024feature,gong2025neuro,gong2025evolutionary,gong2025unsupervised,gong2025agentic}.

\vspace{-0.2cm}
\subsection{Symbolic Equation Extraction Capability}
Symbolic equation extraction (a.k.a., symbolic regression) aims to extract analytical expressions that explain observed input (X) and output (Y) data relationships. 
Traditional approaches, such as genetic programming~\cite{GP,GP2,GP3,GP4,GP5} and its variants with Monte Carlo Tree Search~\cite{sun2022symbolic} or reinforcement learning~\cite{mundhenk2021symbolic}, explore the equation space through iterative search but often face high computational costs, limited scalability, and difficulties in transferring knowledge across datasets. 
Neural and Transformer-based generative models~\cite{biggio2021neural,kamienny2022end} improve efficiency but rely heavily on pretraining data and token-level objectives, which are not perfectly aligned with accuracy, interpretability, and parsimony.
Zero-shot LLM-based approaches~\cite{shojaee2024llm} further demonstrate potential but remain unstable on small or domain-specific datasets.
DataAgents offer a new perspective on equation extraction by transforming this task into a goal-directed, interactive process. Instead of performing one-off static searches, agents can iteratively hypothesize candidate expressions, query symbolic verifiers or numerical solvers for feedback, and refine equations based on data fit and prior reasoning trajectories. With memory, planning, and tool-use capabilities, DataAgents can adapt to new datasets, reuse knowledge from past discoveries, and balance exploration and parsimony, making symbolic regression more scalable, adaptive, and reliable.

\vspace{-0.2cm}
\subsection{Text to SQL Capability}
In text-to-SQL, we can train DataAgents to convert natural language queries into SQL statements by integrating user descriptions with database schemas, enabling non-experts to access structured data efficiently.
Specifically, DataAgents first parses the natural language input to extract intent and entities, and then ground this interpretation to schema elements through techniques such as schema linking (e.g., matching “customer name” to a column cust\_name). 
Next, DataAgents compose the SQL query using syntactic templates or sequence-to-sequence generation models, followed by validation against database constraints. For example, a user request ``Find all orders above \$500 placed in 2023'' is translated into ``SELECT * FROM orders WHERE amount > 500 AND year=2023;''. 
Consider a retail database with tables for orders, customers, and products. An agent can process “Show me the average spending of customers in California” by generating a join between orders and customers filtered by location. Similarly, in a healthcare database, “How many patients were admitted with diabetes in January?” is mapped to a count query on admissions data filtered by diagnosis and date.

\vspace{-0.4cm}
\subsection{Tabular Question Answering Capability}
In tabular QA, DataAgents interpret natural language queries and derive answers from single or multiple structured tables by coordinating tasks like planning and execution. 
For example, DataAgents can process a query like “List employees with salaries above \$50,000 in the HR department.” The agent receives the query, tokenizes it, and parses the intent as a filtering operation (salary > \$50,000, department = HR). It maps “salaries” and “department” to the respective table columns and generates an operation tree for execution.
The steps to achieve tabular QA involve: 
(1) query reception and intent parsing: determine the type of operation (e.g., aggregation, filtering, joining) and the data elements (e.g., specific columns or tables) by mapping  the query to the database schema or tabular structure. 
(2) plan generation using operation trees or chains: 
an operation tree is a hierarchical structure where nodes represent specific data operations (e.g., filtering, joining, aggregating) and edges define the sequence or dependency of these operations.  For example,  a query like “What is the average sales for stores in California?” might result in a tree where the root node is the goal (compute average sales), and child nodes include filtering rows where the state is “California” and aggregating the sales column with a mean operation.
An operation chain is linear sequences of operations. 
(3) data retrieval and augmentation: compute the similarity between query and table schema to identify relevant data, and augment the retrieved tables by incorporating additional context or derived information from LLMs or web sources.
(4) step-by-step execution, 
and (5) result evaluation with refinement. 
In existing studies, AutoTQA~\cite{zhu2024autotqa} employs multi-agent LLMs for multi-table QA, and agents like the Planner create execution plans, while the Executor handles text-to-SQL tasks;
TQAgent~\cite{zhao2025tqagent} integrates knowledge graphs to enhance reasoning over tables and agents like Planner samples operations by confidence scores, and the Worker executes them using pandas dataframes.

\vspace{-0.2cm}
\subsection{Data Quality Assessment Capability}
Data quality assessment refers to the efforts of evaluating the integrity of datasets, including completeness (proportion of non-missing values), accuracy (conformity to true values), consistency (uniformity across entries), and validity (adherence to defined rules). 
This process identifies issues like duplicates or outliers to ensure reliable analysis. 
New requirements arising from large data volumes, continuing assessments in streaming data,  hybrid, heterogeneous, semi-structure formats. 
These required automation beyond manual tools, making AI agents essential for scalable, adaptive processing. 
For instance, Acceldata's Data Quality Agent automates issue detection in pipelines, identifying anomalies like duplicate entries in sales data. Similarly, Alation's agentic AI recommends monitoring checks for enterprise datasets, prioritizing assets based on usage patterns.
Traditionally, analysts manually examine outliers, noises, null ratios; agents use LLMs and tool calling to scan entire datasets, flagging issues like skewed distributions in sales records or inconsistencies in healthcare entries.
Data quality report generation involves: (1) extracting statistical summaries, (3) anomaly detection using rules or tool calling, (4) scoring via metrics like validity rates, and (5) report synthesis. 

\vspace{-0.2cm}
\subsection{Automated Data Repairs Capability}
Automated data repairs aim to identify and correct quality issues in datasets, such as, imputing mean substitutions to replace missing values, removing outliers, merging duplicates, eliminating noises, fixing inconsistent attribute format, avoiding privacy disclosures, to ensure reliability for downstream tasks like machine learning. 
Traditionally, analysts manually profile and correct errors; DataAgents  shift traditional manual scripting and rule-based repairs to autonomous, learning-driven processes that leverages machine learning tools and LLMs to detect and apply fixes proactively.
The key steps of agentic data repairs include: (1)issue detection via tool calling and LLMs token based QA, (2) repair action planning and generation (e.g., imputation), (3) application and validation.

\vspace{-0.2cm}
\section{Call to Actions}
As research on DataAgents remains in its nascent stages, we  discuss several promising research directions in the future.

\vspace{-0.2cm}
\subsection{Open Datasets and Benchmarks for Training DataAgents on Diverse Skills}
Open datasets and benchmarks are essential for training data AI agents on task decompositions, action reasoning, action grounding, and training advanced skills like data preprocessing, data quality assessment, feature engineering, text-to-SQL, symbolic regression, tabular QA, and data repairs. 
There are benchmark datasets for specific skills: tabular QA~\cite{zhong2017seq2sql,yu2018spider,pasupat2015compositional,chen2019tabfact,nan2022fetaqa}, text-to-SQL~\cite{pasupat2015compositional,gan2021towards}, symbolic equation extraction~\cite{matsubara2022rethinking,de2024srbench++}. 
However, there is limited datasets and benchmarks for training agents to learn how to decompose data-related tasks, and other skills like feature engineering, data quality assessment, data repairs.  The research community should curate diverse datasets reflecting real world data-related tasks.

\vspace{-0.2cm}
\subsection{Action Workflow Optimization for DataAgents}
Action workflow optimization involves structuring and ordering decomposed actions, defined as discrete operations reasoned from subtasks within a larger task, in order to maximize efficiency and performance in DataAgents. 
This capability can organize actions into coherent workflows, minimize computational costs, and improves accuracy by prioritizing critical operations. For instance, in tabular QA, we can optimize filtering before aggregation for faster query resolution, or in data repairs, we can prioritize duplicate detection over imputation for consistency.
Typical methods for action workflow optimization include: (1) search-based optimization (e.g., MCTS in AFLOW~\cite{zhang2024aflow}), which explore action combinations, and (2) heuristic-based optimization, which leverages predefined rules. 
Aside from existing studies, we call to develop new solutions to reduce computational overhead in large search spaces and bias in action prioritization, and advance multi-agent collaborative workflow optimization.

\vspace{-0.2cm}
\subsection{Privacy Preservation in DataAgents}
Privacy preservation in DataAgents refers to techniques that protect sensitive information when handling vast datasets, like data preprocessing, quality assessment, and text-to-SQL, thus, is essential to comply with regulations like GDPR~\cite{sharma2019data}. DataAgents process data without exposing personal details, such as preventing leakage of user identities in tabular QA on customer datasets or safeguarding health records in feature engineering for medical analytics.
There are two scenarios that related to privacy preservation. 
First, there is sensitive information in the training data of finetuning DataAgents, and we do not want DataAgents to generate sensitive information as responses to users. 
In this case, appropriate solutions include data editing by detecting sensitive information in reprogram data to finetune DataAgent models, or knowledge editing by machine unlearning to modify DataAgent model parameters to unlearn sensitive information and avoid including sensitive information in generated answers. 
Second, in the post-training QA process, users provide data with sensitive information. In this case, we can add a sensitive information screening step, in order to detect and ignore sensitive information in prompts, or develop a prompt rewriting function to ensure a prompt will not relate DataAgents to sensitive information.

\vspace{-0.2cm}
\subsection{Guardrailing DataAgents Against Malicious Actions}
Action guardrails in data AI agents are mechanisms designed to detect, prevent, or mitigate unauthorized or harmful actions during tasks like tabular QA or text-to-SQL, in order to ensure that outputs align with safety protocols. 
For example, in healthcare, DataAgents should be able to safeguard patient records from unauthorized access during quality assessments
Guardrails are critical to protect against vulnerabilities, such as prompt manipulation leading to data breaches or malicious code execution.
Typical solutions include prompt injection detectors (e.g., classifiers for jailbreaks) which use fine-tuned models to flag direct injections, alignment auditors (chain-of-thought inspectors) which examine reasoning for misalignment,  code analyzers (static analysis engines), action monitoring (real-time inspection and blocking) which employs rule-based filters to halt destructive operations in symbolic regression, and model safety training by enforcing alignment via reinforcement learning from human feedback.

\bibliographystyle{ACM-Reference-Format}
\bibliography{main}
\end{document}